\documentstyle[12pt,psfig,aaai]{article}

\title{SKOPE: A connectionist/symbolic architecture of spoken Korean
processing}
\author{Geunbae Lee \and  Jong-Hyeok Lee \\
Department of Computer Science \& Engineering \\
and Postech Information Research Laboratory \\
Pohang University of Science \& Technology \\
San 31, Hoja-Dong, Pohang, 790-784, Korea \\
gblee@vision.postech.ac.kr
}

\begin{document}

\maketitle

\begin{abstract}
Spoken language processing requires speech and natural language
integration. Moreover, spoken Korean calls for unique processing
methodology due to its linguistic characteristics. This paper presents
SKOPE, a connectionist/symbolic spoken Korean processing engine, which
emphasizes that: 1) connectionist and symbolic techniques must be selectively
applied according to their relative strength and weakness, and  2) the
linguistic
characteristics of Korean must be fully considered for phoneme recognition,
speech and language integration, and morphological/syntactic processing.
The design and implementation of SKOPE demonstrates how connectionist/symbolic
hybrid architectures can be constructed for spoken agglutinative language
processing.  Also SKOPE
presents many novel ideas for speech and language processing. The phoneme
recognition, morphological analysis, and syntactic analysis
experiments show that SKOPE is a viable approach for the spoken Korean
processing.
\end{abstract}

\section{Introduction}
Spoken language processing challenges for integration of speech recognition
into
natural language processing, and must deal with multi-level knowledge sources
from signal level to symbol level. The multi-level knowledge integration and
handling increase the
technical difficulty of both the speech and the natural language processing.
In the speech
recognition side, the recognition must be at phoneme-level for large
vocabulary continuous speech, and the speech recognition module must provide
right level of outputs to the natural language module in the form of not
single solution but many
alternatives of solution hypotheses. The n-best list
\cite{chow:nbest}, word-graph \cite{oerder:word}, and word-lattice
\cite{murveit:large} techniques are mostly used in this purpose.
The speech recognition module can also ask the linguistic
scores from the language processing module in a more tightly coupled
bottom-up/top-down hybrid integration scheme \cite{paul:csr}.
In the natural language side, the insertion, deletion, and
substitution errors of continuous speech must be compensated by robust parsing
and partial parsing techniques, e.g. \cite{baggia:partial}.
Often the spoken languages are ungrammatical, fragmentary, and contain
non-fluencies and speech repairs, and must be processed incrementally under the
time constraints \cite{menzel:parsing}.

Most of the speech and natural language systems which were developed
for English and other Indo-European languages neglect the morphological
processing,
and integrate speech and natural language at the word level
\cite{bates:bbn,agnas:spoken}. Often these systems
employ a pronunciation dictionary for speech recognition and independent
dictionaries for natural language processing. However, for the agglutinative
languages such as Korean and Japanese, the morphological processing plays a
major role in the language processing since these languages have very complex
morphological phenomena and relatively simple syntactic functionality.
Unfortunately
even the Japanese researchers apply degenerated morphological techniques for
the
spoken Japanese processing \cite{hanazawa:atr,sawai:tdnn-lr}. Obviously
degenerated morphological processing limits the usable vocabulary size for the
system,
and word-level dictionary results in exponential explosion in the number of
dictionary
entries. For the agglutinative languages, we need sub-word level integration
which leaves rooms for general morphological processing.

The spoken language processing calls for multi-strategic approaches in order to
deal with signal level as well as symbol level information in a symbiotic and
unified way. Recent development of connectionist speech recognition
\cite{lippman:intro} and connectionist natural language processing
\cite{sharkey:conn} shed lights on the connectionist/symbolic hybrid models of
spoken language processing, and some of
the researches are already available for English and other Indo-European
languages
\cite{wermter:learning,waibel:janus}. We feel that it is the right time to
develop connectionist/symbolic hybrid spoken languages processing systems for
the
agglutinative languages such as Korean and Japanese.

This paper presents one of the such endeavors, SKOPE (Spoken Korean Processing
Engine), that has the following unique features:
1) The connectionist and symbolic techniques are selectively used according
to their strength and weakness. The learning capability, fault-tolerant
property, and ability of simultaneous integration of multiple
signal-level sources make
the connectionist techniques suitable to the phoneme recognition from the
speech
signals, but the structure manipulation and powerful matching (binding)
properties
of the symbolic techniques are the better choices for the complex morphological
processing of Korean. However, the parallel multiple constraint
relaxation capability of the connectionist techniques are applied together with
the symbolic structure binding techniques for the syntactic processing. 2) The
linguistic characteristics of Korean are fully considered in phoneme
recognition, speech and language integration, and morphological/syntactic
processing. 3) The SKOPE provides multi-level application program interfaces
(APIs)
which can utilize the phoneme-level or the morphological level or the syntactic
level services for the applications such as spoken language interface,
voice information retrieval and spoken language translation.

We hope the experience of SKOPE development provide viable answers to
some of the open questions to the speech and language
processing, such as 1) how learning and encoding can be synergetically
combined in speech and language
processing, 2) which aspects of system architecture have to be considered in
spoken language processing, especially in connectionist/symbolic hybrid
systems,
and finally 3) what are the most efficient way of speech and language
integration,
especially for agglutinative languages.

\section{Characteristics of spoken Korean}
\label{sec:char}
This section briefly explains the linguistic characterists of spoken
Korean before describing the SKOPE system. In this paper, Yale romanization is
used for representing the Korean phonemes.
1) A Korean word, called {\em Eojeol}, consists of more than one morphemes with
clear-cut morpheme boundaries.
2) Korean is a postpositional language with many kinds of noun-endings,
verb-endings, and
prefinal verb-endings. These functional morphemes determine the noun's
case roles, verb's tenses, modals, and modification relations between
Eojeols.
3) Korean is a basically SOV language but has relatively free word order
compared to the rigid word-order languages, such
as English, except for the constraints that the verb must appear
in a sentence-final position. However, in Korean, some word-order
constraints do exist
such that the auxiliary verbs representing modalities must follow the main
verb,
and the modifiers must be placed before the word (called head) they modify.
4) The unit of pause in speech (which is called {\em Eonjeol}) may be different
from that of a written text (an Eojeol). The spoken morphological analysis
must deal with
an Eonjeol since no Eojeol boundary can be provided in the speech.
5) Phonological changes can occur in a morpheme, between morphemes in an
Eojeol,
and even between Eojeols in an Eonjeol. These changes include consonant and
vowel assimilation, dissimilation, insertion, deletion, and contraction.
6) Korean has many rising diphthongs that are very similar to mono-vowels at
signal level. Korean has well-developed syllable structures, and unlike
Japanese
that has only CV\footnote{C: consonant, V: vowel} type syllable, Korean has all
different types such as CV, VC, V, CVC. Moreover, in CVC type syllable, first
and second consonants are almost same in pronunciation. These signal
characteristics make it difficult to directly use phonemes or syllables as
sub-word recognition units.

\section{The SKOPE architecture}
\begin{figure}
\centerline{\psfig{figure=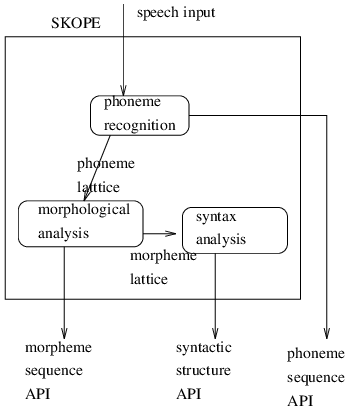}}
\caption{The spoken Korean processing engine architecture. The architecture has
two-level interfaces between modules: phoneme lattice and morpheme lattice for
efficient and generalized speech and natural language integration.}
\label{fg:skope}
\end{figure}

The above spoken Korean characteristics and the relative strength and weakness
of symbolic/connectionist techniques result in the general SKOPE architecture
which is shown in figure~\ref{fg:skope}.
The architecture consists of three different but closely interrelated modules:
phoneme recognition, morphological analysis, and syntactic analysis module.
The phoneme
recognition module processes the signal-level information, and changes it to
the
symbol-level information (phoneme lattice). The morphological analysis begins
the primitive language processing, and connects the speech recognition to the
language processing at the phoneme-level. The syntactic analysis
module finishes the language processing\footnote{We believe that the semantic
and pragmatic processing should be integrated into the domain knowledge for
{\em practical application under the current NLP technology}, so we
excluded the semantic and pragmatic processing from our general model.}, and
produces
the domain independent syntactic structures for application systems.
The following subsections briefly describe each module.

\subsection{Diphone-based connectionist phoneme recognition}
\begin{figure}
\centerline{\psfig{figure=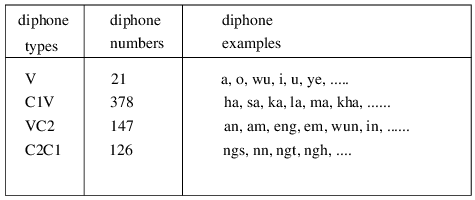}}
\caption{Four different Korean diphone types (V: vowel, C1: syllable-first
consonant, C2: syllable-final consonant)} \label{fg:diphone}
\end{figure}

The phoneme recognition is performed by developing the hierarchically organized
group of
TDNNs (time delay neural networks) \cite{waibel:phoneme}.
Considering the signal characteristics of the Korean phonemes,
we define diphones as a new sub-word recognition
unit. The defined diphones are shown in figure~\ref{fg:diphone},
and are classified into four different types. The diphones
have the co-articulation handling features similar to the
popular triphones \cite{lee:auto} but are much fewer in numbers.

\begin{figure}
\centerline{\psfig{figure=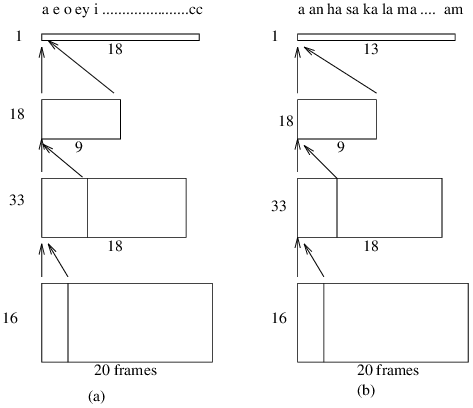}}
\caption{(a) The TDNN architecture for the vowel group identification. Note the
cc group contains no vowels. (b) The architecture of sub-TDNNs for /a/ vowel
group. The other 17 sub-TDNNs have the same architecture, but different number
of output units according to the number of diphones in each of the vowel
group.}
\label{fg:dtdnn}
\end{figure}

Figure~\ref{fg:dtdnn} shows the
architecture of the component TDNNs in the phoneme recognition module.
The whole module consists of total 19
different TDNNs for recognition of the defined Korean diphones. The top-level
TDNN identifies the 18 vowel groups of diphones (we re-classified the total 672
diphones into 18 different groups according to the vowels that are contained in
the diphones). The 18 different sub-TDNNs recognize the target diphones.

For the training of TDNNs, we manually segment the digitized speech into 200
msec range (which includes roughly left-context phoneme, target diphone, and
right context phoneme), and perform 512 order FFTs and 16 step mel-scaling
\cite{waibel:phoneme} to get the filter-bank coefficients. Each frame size is
10
msec, so 20 (frames) by 16 (mel-scaling factor) values are fed to the TDNNs
with
the proper output symbols, that is, vowel group name or target diphone names.
After the training of each TDNN, the phoneme recognition is performed by
feeding 200 msec signals to the vowel group identification network and
subsequently to the
proper diphone recognition network. The 200 msec signals are shifted by 30 msec
steps and continuously fed to the networks to process the continuous speech
in an Eonjeol.  From the resulting diphone
sequences, the necessary phoneme lattice has to be constructed.
We use a simple deterministic decoding heuristics and try to maintain all the
possible diphone spotting results since the later
phonological/morphological processing can safely prune the incorrect
recognitions.  The decoding begins by grouping the diphones into the same
types (see figure~\ref{fg:diphone}). The frequency count for each diphone, that
is,
the number of specific diphones per 30 msec frame shift, is utilized to
fix the insertion errors by deleting the lower frequency count diphones,
and finally the diphones are split into the constituent phonemes by
merging the same phonemes in the neighboring diphones.

\subsection{Table-driven morphological and phonological analysis}
The morphological analysis starts with the phoneme lattice.
The phoneme lattice delivers the alternative phonetic
transcriptions\footnote{Unlike English, the Korean alphabet is truly phonetic
in the sense that each phoneme is pronounced as it
is written. That is why we sometimes use {\em phonetic} and
{\em phonemic} interchangeably.} of input speech, which
must be searched by the morphological/phonological
analyzer to reconstruct the orthographic morpheme strings. The conventional
morphological analysis procedure \cite{sproat:morph}, that is, morpheme
segmentation,
morphotactics modeling, and orthographic rule (or phonological rule) modeling,
must be augmented and extended as the followings: 1) The conventional
morpheme segmentation is extended
to deal with the exponential number of phoneme sequences and
between-morpheme phonological changes during the segmentation, 2) the
morphotactics modeling is extended to cope with the complex verb and
noun-endings
(or postpositions), and 3) the orthographic rule modeling is combined with
the phonological rule modeling to correctly transform the phonetic
transcriptions to the orthographic morpheme sequences.

\begin{figure}
\centerline{\psfig{figure=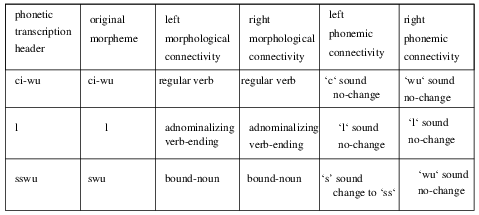}}
\caption{The morpheme-level phonetic dictionary.}
\label{fg:dict}
\end{figure}

The central part of the morphological analysis lies in the dictionary
construction. In our dictionary, each phonetic transcription of single morpheme
has a separate dictionary entry. Figure~\ref{fg:dict} shows the unified
dictionary both for speech and language processing (called morpheme-level
phonetic dictionary) with three different morpheme entries {\em ci-wu, l, swu}.

\begin{figure}
\centerline{\psfig{figure=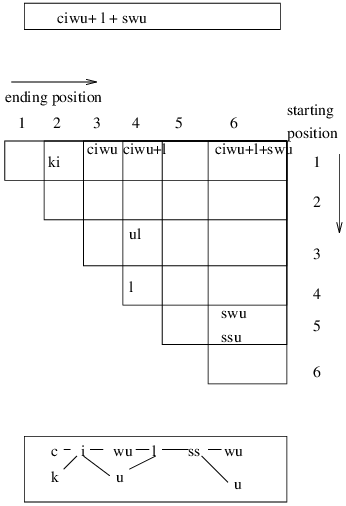}}
\caption{Morphological parsing of the phoneme lattice (from top:
output morpheme sequence in an Eonjeol, triangular parsing table, input phoneme
lattice).}
\label{fg:morph}
\end{figure}

The extended morphological analysis is based on the well-known tabular
parsing technique for context-free language \cite{aho:theory} and augmented to
handle the Korean
phonological rules and phoneme-lattice input. Figure~\ref{fg:morph} shows our
extended table-driven morphological analysis process. The example phoneme
lattice was obtained from the input speech
{\it ci-wul-sswu} (removable), and the morphological analysis
produces {\it ci-wu+l+swu} (remove+ADNOMINAL+BOUND-NOUN),
where '+' is the morpheme boundary, and '-' is the syllable boundary.

The extended morpheme segmentation is basically performed using the
dictionary search. During
the left-to-right scan of the input phoneme lattice, when a morpheme boundary
is
found
in the lattice, the morpheme is enrolled in the triangular table in an
appropriate
position. For example,
in figure~\ref{fg:morph}, morphemes such as {\em ci-wu, l, swu, etc} are
enrolled in the table position (1,3), (4,4), (5,6), etc. The position (i,j)
designates the starting and ending position of the enrolled morphemes.
However since the
input is a phoneme-lattice, total exponential time is required to
find all the possible morpheme boundaries. To cope with such exponential
explosion, the dictionary is organized as trie
structure \cite{aho:data} using the phonetic transcriptions as trie indices,
and breadth-first search of the trie can prune the unnecessary phoneme
sequences earlier in the search.

The morphotactics modeling is necessary after all the morphemes are enrolled in
the table in order to combine only legal morphemes into an Eojeol (Korean
word),
and the process is called morpheme-connectivity-checking.
Since Korean has well developed postpositions (noun-ending, verb-ending,
prefinal verb-ending) which play as grammatical functional morphemes, we must
assign each morpheme proper part-of-speech (POS) tags for the efficient
connectivity
checking. Our more than 200 POS tags which are refined from the 13 major
Korean lexical categories are
hierarchically organized, and contained in the dictionary
(in the name of morphological connectivity, see figure~\ref{fg:dict}).
In the case of idiomatic
expressions, we place such idioms directly in the dictionary for efficiency,
where two different POS tags are necessary for the left and the right
morphological connectivity.
For single morpheme, the left and the right POS tags are always same.
The separate morpheme-connectivity-matrix indicates
the legal morpheme combinations, and the morphotactics modeling
is performed using the POS tags (in the dictionary) and
morpheme-connectivity-matrix.

The orthographic rule modeling must be integrated with the phonological rule
modeling in spoken language processing. Since we must deal with the phoneme
lattice, the conventional rule-based modeling requires exponential number of
rule application \cite{koskenniemi:two}. So our solution is based on the
declarative modeling of both orthographic and phonological rules in uniform
way.
That is, in our dictionary, the conjugated verb forms as well as the original
verb forms are enrolled, and the same morphological connectivity information is
applied.
The phonological rule modeling is also accomplished declaratively by having
the phonemic
connectivity information in the dictionary. The phonemic connectivity
information for each morpheme declares the possible phonemic changes in the
first
(left) and last (right) positioned phonemes in the morpheme, and the separate
phoneme-connectivity-matrix indicates the legal sound combinations in Korean
phonology. For example, in figure~\ref{fg:morph}, the morpheme {\em l} can be
combined with the morpheme {\em swu} during the morpheme connectivity checking
even if {\em swu} is actually pronounced as {\em sswu} because
the phoneme-connectivity-matrix supports the legality of the
combination of {\em l} sound with {\em ss} sound\footnote{This legality comes
from the Korean phonology rule {\em glotalization} (one form of consonant
dissimilation) stating that {\em s} sound becomes {\em ss} sound after
{\em l} sound.}.  In this way, we can declaratively model all the major Korean
phonology rules such as second consonant standardization, consonant
assimilation,
palatalization, glotalization, insertion, deletion, and contraction.

\subsection{Table-driven connectionist/symbolic syntax analysis}
The phoneme lattice-based morphological analysis produces the morphologically
analyzed (segmented and stem reconstructed) morpheme sequences. Since there are
usually more than one analysis results due to the errors of speech recognition
process, the outputs are usually organized as morpheme lattice.
For the seamless integration of the
morphological analysis with the syntax analysis, we employ the same
table-driven
control for the syntax analysis as well as the morphological analysis.

We extend the category formation and functional application rules in the
previous categorial unification
grammar\cite{zeevat:combining,uszkoreit:categorial}
to deal with the word order variations in Korean:
\begin{itemize}
\item if category a $\in$ C, then a $\in$ C'
\item if category a $\in$ C', and category set S $\in$ C', then a/S $\in$ C'
and
a$\setminus$S $\in$ C'
\end{itemize}
where S is an unordered set of categories.
\begin{itemize}
\item left cancellation: $a_i$ b$\setminus$\{$a_1$,$a_2$, \ldots, $a_n$\}
results in
b$\setminus$\{$a_1$, $a_2$, \ldots, $a_{i-1}$, $a_{i+1}$, \ldots, $a_n$\}
\item right cancellation: b/\{$a_1$,$a_2$, \ldots, $a_n$\} $a_i$ results in
b/\{$a_1$, $a_2$, \ldots, $a_{i-1}$, $a_{i+1}$, \ldots, $a_n$\}
\end{itemize}

\begin{figure}
\centerline{\psfig{figure=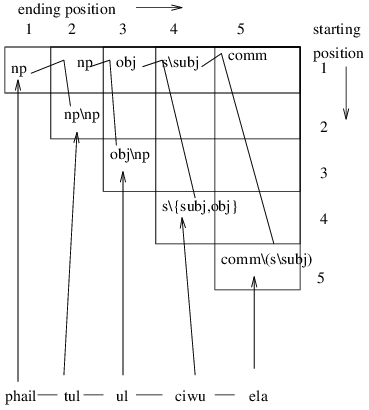}}
\caption{Table-driven interactive relaxation parsing of a categorial grammar.
The input sentence is {\em phai-l-tul-ul ciwu-ela} (delete the files).
Only single morpheme chain and only one sense for each morpheme
is shown as input for clear illustration. The {\em subj, obj, comm}
indicate {\em np[subj], np[obj], s[command]}, respectively. The table contains
only the nodes that participate in the final parse trees.}
\label{fg:parsing}
\end{figure}

The syntax analysis is performed by interactive relaxation (spreading
activation)
parsing on the categorial grammar where the position of the functional
applications are controlled by a triangular table. The original interactive
relaxation parsing
\cite{howells:conn} was extended to provide efficient constituent searching and
expectation generation through positional information provided by categorical
grammar and triangular table. Figure~\ref{fg:parsing} shows table-driven
interactive relaxation parsing.

The interactive relaxation process consists of the following three steps that
are repetitively executed: 1) add nodes, 2) spread activation, and 3) decay.
\begin{description}
\item[add nodes] Grammar nodes (syntactic categories from the dictionary)
are added for each sense of the morphemes when
the parsing begins. A grammar node which has more activation than the
predefined threshold $\Theta$ generates new nodes in the proper positions (to
be
discussed shortly). The newly
generated nodes search for the constituents (expectations)
which are in the appropriate table
positions, and are of proper function applicable categories. For example,
in figure~\ref{fg:parsing}, when np$\setminus$np(2,2) fires, it generates
np(1,2). The
generated np(1,2) searches for the constituents np(1,1) to be combined with
np$\setminus$np(2,2).
\item[spread activation] The bottom-up spreading activation is as follows:\\
$n \times \rho \times a \times \frac{{a_i}^2}{\sum_{j=1}^{n} {a_j}^2}$\\
where predefined portion $\rho$ of total activation $a$ is passed upward to the
node with activation $a_i$ among the $n$ parents each with node activation
$a_j$. In other words, the node with
large activation gets more and more activation, and it gives an inhibition
effects without explicit inhibitory links \cite{reggia:properties}.
The top-down spreading activation uniformly distributes:\\
$\rho' \times a$\\
among the children where $\rho$' is predefined portion of the source
activation $a$.
\item[decay] The node's activation is decayed with time. The node with less
constituents than needed gets penalties plus decays:\\
$a \times (1-d) \times \frac{Ca}{Cr}$\\
where $a$ is an activation value, $d$ is a decay ratio, and Ca, Cr is the
actual
and required constituents. After the decay, the node with less activation than
the predefined threshold $\Phi$ is removed from the table.
\end{description}

The node generation and constituent search positions are controlled by the
triangular table. When the node a(i,j) acts as an argument, it generates node
only in the position (k,j) where $1<k<j$, and the generated node searches
for the constituents
(functors) only in the position (k,i-1). Or when the node is generated in the
position
(i,k) where $j<k<number-of-morphemes$, it searches for the position (j+1,k)
for its constituents.
When the node acts as a functor, the same position restrictions also
apply for the node generation and the argument searching.
The position control combined with the interactive relaxation guarantees an
efficient,
lexically oriented, and robust syntax analysis of spoken languages.

\section{Implementation and experiments}
The SKOPE was fully implemented in UNIX/C platform, and have been extensively
tested in practical domains such as natural language interface to operating
systems.
The phoneme recognition module targets 1000 morpheme continuous speech,
currently speaker dependent due to the short of standard speech database for
Korean. The unified morpheme-level phonetic dictionary has about 1000
morpheme entries and compiled into the trie structure. The
morpheme-connectivity-matrix and phoneme-connectivity-matrix are encoded with
the special Korean POS (part-of-speech) symbols and compressed.

\begin{figure}
\centerline{\psfig{figure=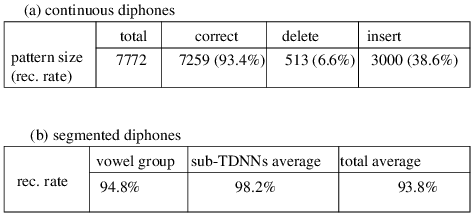}}
\caption{(a) Continuous diphone recognition versus (b) segmented diphone
recognition}
\label{fg:res0}
\end{figure}

This section demonstrates the SKOPE's performance in continuous diphone
recognition, morphological analysis, and syntax analysis experiments.
For the continuous diphone recognition experiment, we generated about 5500
diphone patterns from the 990 Eojeol patterns (66 Eojeols, 15 times
pronunciation) for the training of TDNNs.
In the performance phase, the new 2600 test Eojeol patterns (260 Eojeol, 10
times pronunciation) are
continuously shifted with 30 msec step, and generate 7772 test
diphone patterns disjoint from the training patterns.
Figure~\ref{fg:res0}-a shows the continuous diphone
recognition performance.  The {\em correct} designates that the correct
target diphones were spotted in the testing position, and the {\em delete}
designates the other case.
The {\em insert} designates that the non-target diphones were spotted in the
testing position. To compare the ability of handling the continuous speech,
we also tested the diphone recognition using the hand-segmented test patterns
with the same 7772 target diphones.
Figure~\ref{fg:res0}-b shows the segmented diphone recognition performance.
Since the test data are already hand-segmented before input, there are no
insertion and deletion errors in this case.
The fact that the segmented speech performance is not much better than the
continuous one (93.8\% vs. 93.4\%) demonstrates the
diphone's suitability to handling the continuous speech.

\begin{figure}
\centerline{\psfig{figure=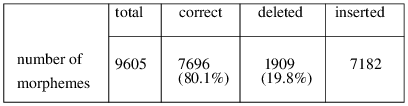}}
\caption{Morphological analysis from continuous speech signals. The table
indicates that, among the total 9605 morphemes in 2600 Eojeol patterns, the
80.1\% are correctly recognized and analyzed, and 19.8\% cannot be analyzed for
deletion errors. The 7182 spurious morphemes are also generated due to the
speech
insertion errors.}
\label{fg:res1}
\end{figure}

For the morphological analysis performance, we used the same
990 Eojeol patterns to train the phoneme recognition module, and the
2600 Eojeol patterns to test the morphological analysis performance directly
from the speech input.
Figure~\ref{fg:res1} shows the results.

\begin{figure}
\centerline{\psfig{figure=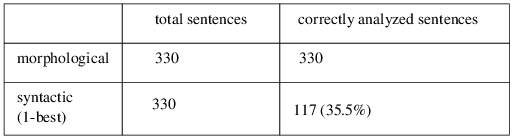}}
\caption{The morphological and syntactic analysis from the artificially made
phoneme lattices.}
\label{fg:res2}
\end{figure}

This experiment shows that most of the morphological errors are
propagated from the incorrect (deleted) or spurious (inserted) phoneme
recognition results.
To see the original performance of the morphological and syntactic analysis
modules assuming no speech recognition error, we
artificially made the phoneme lattices by mutating the correctly
recognized phoneme sequences
according to the phoneme recognizer's confusion matrix. Each phoneme
lattice was made to contain at least one correct recognition result,
so the phoneme recognition performance is assumed to be perfect except the
artificially made insertion errors (mutations). In this way, we
made 6 or 7 lattices for each of the 50 sentences, altogether 330
phoneme lattices.  The average phoneme alternatives per single correct phoneme
in the lattice are 2.3,
and average sentence length is 31 phonemes. This means there are average
$2.3^{31}$ phoneme chains in each lattice. The used sentences are natural
language commands to UNIX \cite{lee:from} and are fairly complex which have
one or two embedded sentences or conjunctions.
Figure~\ref{fg:res2} shows the morphological and syntactic analysis results for
these artificially made phoneme lattices. For
the syntactic level interactive relaxation, we used the following parameters
(which are experimentally
determined): upward propagation portion $\rho$ 0.05,
downward propagation portion $\rho$' 0.03, decay ratio $d$ 0.87, the node
generation threshold $\Theta$ 0.51, and the node removal threshold $\Phi$
0.066.

The morphological analysis was perfect as shown in the table. Since the
phoneme lattice was made to contain at least one correct phoneme recognition
result, the morphological analysis must be perfect as long as the morpheme is
enrolled in the dictionary and the connectivity information can cover all
the morpheme combinations. This was possible due to the small number of
tested sentences (50 sentences). This results verify that most of the
morphological
analysis errors from real speech input are actually propagated from the phoneme
recognition errors as discussed before. However, the syntax analysis results
are
marginal here since we only count the single best scored tree, and we don't use
yet any semantic feature in the analysis.
The syntax analysis failures mainly come from 1) the
insertion errors (artificial mutations) in the phoneme
lattices\footnote{Recall we
generated average 2.3 phonemes per single correct phoneme.}, which result in
ambiguous
morpheme lattice, and finally produce redundant syntax trees, and 2)
the inherent structural ambiguities in the sentence. These failures
should be greatly reduced if we generate n-best scored parse trees, and let
the semantic processing module select the correct ones as is usually done
in most of the probabilistic parsing schemes\cite{charniak:stat}.

\section{Conclusions and future works}
This paper explains the design and implementation of spoken Korean processing
engine, which is a connectionist/symbolic hybrid model of spoken language
processing by utilizing the linguistic characteristics of Korean.
The SKOPE model demonstrates the synergetic integration of connectionist
and symbolic techniques by considering the relative strength and weakness of
two
different techniques, and also demonstrates the phoneme level speech and
language integration for
general morphological processing for agglutinative languages. Besides the
above two
major contributions, the SKOPE architecture has the following unique features
in spoken language processing: 1) the diphones are newly developed as a
sub-word
recognition unit for connectionist Korean speech recognition, 2) the
morphological and syntactic analysis are tightly coupled by using the uniform
table-driven control, 3) the phonological and orthographic rules are
uniformly co-modeled declaratively, and 4) the table-driven interactive
relaxation parsing and extension of the categorial grammar can provide robust
handing of word-order variations in Korean.

However, current implementation of the system still suffers
from excessive continuous speech recognition errors. Since the large vocabulary
continuous
speech recognition is still an open problem, we cannot hope for the 100\%
correct speech recognition results in the near future. Currently, we are
pursuing multi-strategic approaches to the advanced spoken language
processing model, including optimizing TDNN-based phoneme recognition
module, integrating HMM-based morpheme recognition module into the
connectionist phoneme recognition, and incorporating
probabilistic searches into the morphological analysis process as well as the
syntactic analysis process. We are also developing applications on top of our
SKOPE, including speech-to-speech translation system and intelligent interface
agent for UNIX operating system. We hope our approach could be extended to
other
agglutinative languages such as Japanese, Finish, and Turkish, and also to the
languages that have complex morphological phenomena such as German and Dutch.

\section*{Acknowledgments}
This research was supported partly by a grant from KOSEF (Korea Science
and Engineering Foundation) and PIRL (Postech Information Research
Laboratory).  The SKOPE's various modules were programmed by our students: the
phoneme recognition module by Kyunghee Kim \& Kyubong Baac, the morphological
analysis module by EunChul Lee \& Wonil Lee, and finally the syntax analysis
module by Wonil Lee.

\bibliography{/mnt/home/1/HCI/gblee/paper/nlpsp}
\bibliographystyle{aaai}
\end{document}